\documentclass[11pt,a4paper]{article}
\usepackage[T1]{fontenc}
\usepackage[utf8]{inputenc}
\usepackage{lmodern}
\usepackage[british]{babel}
\usepackage[a4paper,margin=2cm,includefoot,footskip=1cm]{geometry}
\usepackage{amsmath,amssymb,mathtools}
\usepackage{graphicx,adjustbox}
\usepackage{array,longtable,booktabs,calc}
\usepackage{caption,needspace,microtype}
\usepackage{titlesec}
\usepackage[round,authoryear]{natbib}
\usepackage{xurl}
\usepackage[hidelinks]{hyperref}
\DeclareUnicodeCharacter{0394}{\ensuremath{\Delta}}

\counterwithin{table}{section}
\counterwithin{figure}{section}

\titleformat{\section}{\normalfont\large\bfseries}{\thesection}{0.65em}{}
\titleformat{\subsection}{\normalfont\normalsize\bfseries}{\thesubsection}{0.65em}{}
\titlespacing*{\section}{0pt}{2.0ex plus .5ex minus .2ex}{0.8ex}
\titlespacing*{\subsection}{0pt}{1.5ex plus .3ex minus .2ex}{0.5ex}
\hypersetup{pdftitle={Improving the Intrinsic Interpretability of Differentiable Decision Trees through Euclidean Feature-Concentration Regularisation},pdfauthor={Qiqi Yang},pdfsubject={Machine Learning; Differentiable Decision Trees},pdfkeywords={EU-DDT, interpretable machine learning, differentiable decision trees, reinforcement learning}}
\title{\LARGE\bfseries Improving the Intrinsic Interpretability of Differentiable Decision Trees through Euclidean Feature-Concentration Regularisation}
\author{Qiqi Yang\thanks{This work is based on the author's MSc Applied Mathematics dissertation at the University of Birmingham.}}
\date{}
\begin{document}
\maketitle
\begin{abstract}
Differentiable decision trees (DDTs) support end-to-end optimisation and explicit
decision paths, but dense oblique splits can obscure the role of individual input
features. We propose the Euclidean-Concentrated Differentiable Decision Tree (EU-DDT),
which regularises the relative absolute feature weights at each internal node using
their normalised squared Euclidean distance from the uniform distribution. This
encourages dominance by a small number of features while retaining probabilistic soft
routing. Across five matched random seeds, EU-DDT reduces the mean effective number of
features from 8.94 to 1.54 on Wine and from 45.28 to 9.06 on Digits. Mean test accuracy
changes from 100.0\% to 93.9\% on Wine and from 92.3\% to 94.6\% on Digits, illustrating
the dataset-dependent trade-off between concentration and predictive performance. We
also use EU-DDT to parameterise a PPO actor with an MLP critic. Two random-seed runs per
environment show a consistent qualitative increase in concentration while preserving
effective task performance. In representative CartPole-v1 and MiniPacman runs, EU-DDT
reaches 100\% final evaluation success, while mean node concentration increases from
0.1272 to 0.7179 and from 0.0437 to 0.7237, respectively. These results support EU
regularisation as a mechanism for reducing node-level feature complexity; the
reinforcement-learning evidence remains preliminary because only two seeds are
evaluated.
\end{abstract}

\Needspace{6\baselineskip}\section{Introduction}

\subsection{Research Background and Motivation}

Conventional decision trees use Boolean hard splits, which yield
explicit node rules but make end-to-end gradient optimisation of the
whole tree difficult. A DDT uses sigmoid gates to generate continuous
branch probabilities, allowing internal nodes, path probabilities, and
leaf parameters to be optimised jointly by backpropagation \citep{frosst2017soft}. Its fixed tree topology and explicit root-to-leaf paths
provide structural transparency. Nevertheless, the dense weight vector
of an oblique soft split can still make a single node depend
simultaneously on many input dimensions. This paper therefore
retains probabilistic soft routing while directly constraining the
allocation of relative feature weights within each node, combining
optimisability and explicit paths with node decisions dominated by a
small number of features.

In this paper, intrinsic structural interpretability is
operationalised as the combination of explicit root-to-leaf decision
paths and low-dimensional, dominant-feature dependence at individual
internal nodes.

\subsection{Research Questions}

To assess whether EU-DDT can improve the intrinsic structural
interpretability of a DDT by increasing feature concentration within its
internal nodes while preserving task performance, this paper
addresses three related research questions concerning the regularisation
mechanism, supervised classification performance, and PPO policy
representation.

RQ1: To what extent can normalised squared Euclidean concentration
regularisation increase the concentration of relative feature weights
within DDT internal nodes and encourage node representations dominated
by a small number of features?

RQ2: Under identical DDT architectures and training conditions, how does
EU-DDT compare with an unregularised DDT, L1-DDT, and L2-DDT in
classification accuracy and node-concentration metrics; and, when
neural-network accuracy is used as an external performance reference,
what balance does EU-DDT achieve between predictive performance and
intrinsic structural interpretability?

RQ3: When EU-DDT is used to parameterise a PPO actor, can the EU
concentration mechanism developed for supervised learning increase the
concentration of the actor\textquotesingle s dependence on a small
number of state features while maintaining effective policy performance?

\subsection{Research Method and Main Contributions}

The central contribution of this paper is a continuous Euclidean
concentration regulariser that acts directly on the normalised weight
distribution within each DDT node. The method leaves the baseline
DDT\textquotesingle s sigmoid probabilistic routing and end-to-end
optimisation unchanged, while shifting node decisions from diffuse
dependence on many features towards dominance by a small number of
features. Five-seed supervised comparisons evaluate performance
preservation and concentration gains, and the same mechanism is extended
to a PPO actor. Two independent random-seed repetitions further assess
whether the direction of the reinforcement-learning findings is
consistent.

\Needspace{6\baselineskip}\section{Background and Related Work}

\subsection{Differentiable Decision Trees and Node Interpretability}

Conventional CART uses axis-aligned hard splits, making root-to-leaf
paths directly readable, but its discrete splits are not amenable to
global gradient-based optimisation \citep{breiman1984cart}. Neural
decision forests subsequently connected representation learning with
tree models through neural split functions \citep{rotabulo2014ndf}, and developed into deep neural decision forests in which split
and leaf parameters can be optimised jointly \citep{kontschieder2015dndf}. Soft decision trees further use sigmoid gates to define branch
probabilities and root-to-leaf paths explicitly, allowing all parameters
to be updated jointly by gradient descent \citep{frosst2017soft}. In
accordance with the main implementation, this paper forms a
path-probability-weighted mixture of leaf logits and then applies
Softmax to obtain class probabilities.

\subsection{Feature Selection, Regularisation, and Research Positioning}

An oblique DDT node defines a soft split by a multidimensional
hyperplane. DNDT implements a decision tree as a neural architecture
that can be trained end to end by gradient descent and discusses the
self-pruning behaviour that emerges during training \citep{yang2018dndt}.
Adaptive Neural Trees additionally learn tree growth and module
configuration \citep{tanno2019ant}. For feature selection, NODE uses
entmax to produce exactly zero weights \citep{popov2020node}, whereas
ICCT uses differentiable crispification to learn few-feature or
single-feature splits and notes that L1 constraints typically produce
near-zero rather than strictly sparse rules \citep{paleja2022icct}.
EU-DDT applies neither hard argmax nor a sparse selection operator.
Instead, it retains sigmoid-based oblique soft routing and directly
optimises the concentration of relative weights within each node,
thereby improving the intrinsic structural interpretability of the model
that actually makes the predictions.

\subsection{Interpretable Reinforcement Learning}

Differentiable decision trees can serve as policy functions and be
optimised directly by policy gradients \citep{silva2020ddtrl,silva2021prolonets}. Natural-language policy specifications can also be
converted into lexical decision trees to initialise reinforcement
learning, demonstrating that tree policies can carry human-facing prior
knowledge and provide an inspectable interface \citep{tambwekar2023nl}.
ICCT learns sparse tree policies through differentiable crispification,
whereas DTPO directly optimises hard decision-tree policies \citep{paleja2022icct,vos2024dtpo}. In this paper, EU-DDT
parameterises the actor and is updated by PPO; the MLP critic is used
only for value estimation and advantage construction. The tree topology,
root-to-leaf paths, and dominant state variables therefore form an
inspectable representation of the policy that is actually deployed.

\Needspace{6\baselineskip}\section{Baseline Differentiable Decision Tree Model}

\subsection{Soft Routing at Internal Nodes}

The four DDT variants share the same complete binary-tree topology, leaf
representation, and prediction procedure; they differ only in the
regularisation term used in the training objective. For sample \(x_{b}\)
and internal node \(i\), the node uses a weight vector \(w_{i}\),
threshold \(c_{i}\), and inverse temperature \(\alpha\) to define a
sigmoid gating probability. Under the path convention used in this
paper, the probability of taking the left branch is \(g_{bi}\),
and that of taking the right branch is \(1 - g_{bi}\).

\[z_{bi} = \alpha h_{bi} = \alpha(w_{i}^{{\top}}x_{b} - c_{i}).\]

The scaled distance from the node hyperplane and the corresponding
sigmoid gating probability are defined as

\[g_{bi} = \sigma(z_{bi}) = \frac{1}{1 + exp( - z_{bi})}.\]

The baseline DDT split is determined by \(w_{i}^{\top}\ x_{b}\) and is
therefore an oblique soft split that may depend simultaneously on
multiple input dimensions. EU-DDT does not alter this splitting
function. Instead, during training it encourages the hyperplane normal
vector \(w_{i}\) to be dominated by a small number of input dimensions,
thereby reducing the feature-attribution complexity of each node
decision.

\subsection{Path Probabilities and Leaf Outputs}

For leaf \(\ell\), let \(L_{\ell}\) denote the set of internal nodes at
which the sigmoid branch is selected along the path from the root to
\(\ell\),

and let \(R_{\ell}\) denote the set of internal nodes at which the
complementary branch is selected along the same path. Because the
probabilities of the two branches sum to one, they form a complete
probabilistic routing mechanism at every node.

For sample \(x_{b}\), the probability of reaching leaf \(\ell\) is
therefore

\[\pi_{b\ell} = \prod_{i \in \text{L}_{\ell}}^{}g_{bi}\prod_{i \in \text{R}_{\ell}}^{}{(1 - g_{bi})}.\]

\subsection{Predictive Distribution}

Each leaf \(\ell\) stores a learnable vector of class logits
\(a_{\ell}\). The implementation first forms a path-probability-weighted
mixture of the leaf logits and then applies Softmax to the aggregated
logit vector:

\[{\overset{\hat{}}{p}}_{b} = Softmax(\sum_{\ell = 1}^{L}{\pi_{b\ell}a_{\ell})}\]

\subsection{Classification Loss}

Supervised classification uses the categorical cross-entropy between the
final predictive distribution and the one-hot label:

\[\text{L}_{\text{CE}} = - \frac{1}{B}\sum_{b = 1}^{B}{\sum_{c = 1}^{K}{y_{bc}log{\overset{\hat{}}{p}}_{bc}.}}\]

\subsection{Training Objectives and Comparison Models}

The internal-node parameters and leaf logits are optimised jointly under
the same objective. The objectives for the unregularised,
L1-regularised, L2-regularised, and EU-DDT models are, respectively,

\[\text{L}_{\text{baseline}} = \text{L}_{\text{CE}}\]

\[\text{L}_{\text{L}1} = \text{L}_{\text{CE}} + \lambda_{1}R_{\text{L}1},\]

\[\text{L}_{\text{L}2} = \text{L}_{\text{CE}} + \lambda_{2}R_{\text{L}2},\]

EU-DDT uses

\[\text{L}_{\text{EU}} = \text{L}_{\text{CE}} + \lambda_{\text{EU}}R_{\text{EU}}.\]

All four models share the same routing mechanism, leaf representation,
initialisation, optimiser, and number of training epochs. They differ
only in the regularisation objective applied to the internal-node
weights.

L1 and L2 constrain parameter magnitude, whereas EU regularisation
constrains the concentration of the normalised relative weights.

\Needspace{6\baselineskip}\section{EU-DDT and Normalised Squared Euclidean Concentration Regularisation}

\subsection{Normalised Weights and Concentration}

EU-DDT retains the sigmoid-based oblique routing defined in Section 3
and applies a continuous concentration constraint only to the relative
feature weights of the internal nodes. For node i, the absolute weights
are first normalised into the distribution

\[\rho_{ij} = \frac{|w_{ij}| + \epsilon}{\sum_{k = 1}^{d}{(|w_{ik}| + \epsilon)}}.\]

\[C_{i} = \frac{\sum_{j = 1}^{d}{(\rho_{ij} - \frac{1}{d})}^{2}}{1 - \frac{1}{d}}.\]

\(C_{i}\) is the normalised squared Euclidean distance between the
node\textquotesingle s normalised weight distribution and the uniform
distribution. When all feature shares are equal, \(C_{i}\) = 0; when one
feature holds all of the mass in the limiting case, \(C_{i}\) = 1. This
quantity therefore directly measures the extent to which a node decision
is dominated by a small number of features.

\subsection{Mathematical Properties and Training Objective}

The normalised weights \(\rho_i\) lie on the probability simplex
\(\Delta^{d-1}\subset\mathbb{R}^{d}\). Expanding the squared distance
from the uniform distribution gives the identity

\[\boxed{\sum_{j=1}^{d}\left(\rho_{ij}-\frac{1}{d}\right)^2
=\sum_{j=1}^{d}\rho_{ij}^{2}-\frac{1}{d}.}\]

Consequently,

\[C_i=\frac{\sum_{j=1}^{d}\rho_{ij}^{2}-\frac{1}{d}}
{1-\frac{1}{d}}.\]

The uniform distribution and the single-feature limit are,
respectively, the minimum- and maximum-concentration cases. Hence,

\[0 \leq C_{i} \leq 1.\]

\subsection{Effective Number of Features and Regularised Objective}

For an intuitive interpretation of node complexity, the effective number
of features is defined as

\[N_{\text{eff},i} = \frac{1}{\sum_{j = 1}^{d}\rho_{ij}^{2}}.\]

\[N_{\text{eff},i} = \frac{d}{1 + (d - 1)C_{i}}.\]

Because this relationship is nonlinear, the mean effective number of
features is not obtained by substituting the mean concentration into the
above expression:

\[\overline{N}_{\mathrm{eff}} \neq
\frac{d}{1+(d-1)\overline{C}}.\]

For a tree with \(N\) internal nodes, the mean node concentration is

\[\overline{C}=\frac{1}{N}\sum_{i=1}^{N}C_i.\]

\Needspace{4\baselineskip}Since optimisation minimises the training objective, the EU
regularisation term is defined as

\[R_{\mathrm{EU}}=1-\overline{C}.\]

The complete supervised-learning objective is

\[\mathcal{L}_{\mathrm{EU-DDT}}=\mathcal{L}_{\mathrm{CE}}
+\lambda_{\mathrm{EU}}R_{\mathrm{EU}}.\]

\(\lambda_{EU}\) controls the relative weighting of the classification
objective and the node-concentration objective. When \(\lambda_{EU}\) =
0, the model reduces to the unregularised DDT. As \(\lambda_{EU}\)
increases, optimisation places a stronger preference on node structures
dominated by a small number of features, while retaining continuous soft
splits rather than forcing a single non-zero feature.

The objective therefore jointly pursues task performance and node-wise
feature concentration within a single differentiable optimisation
process. The experiments test whether the selected coefficient produces
a clear concentration gain while preserving accuracy.

\subsection{Distinction from L1 and L2 Regularisation}

L1 and L2 serve as two baselines that distinguish parameter-magnitude
shrinkage from relative feature concentration within a node. The
experiments use the following regularisers, averaged over the numbers of
nodes and feature dimensions:

\[R_{\text{L}1} = \frac{1}{Nd}\sum_{i = 1}^{N}{\sum_{j = 1}^{d}{|w_{ij}|.}}\]

\[R_{\text{L}2} = \frac{1}{Nd}\sum_{i = 1}^{N}{\sum_{j = 1}^{d}{w_{ij}^{2}.}}\]

L1 directly penalises the absolute values of the weights and can
encourage near-zero parameters, whereas L2 penalises squared magnitudes
and shrinks the weight vector as a whole. Both operate on the scale of
the raw parameters. EU regularisation first normalises the absolute
weights of each node and then directly optimises the concentration of
the resulting feature shares relative to the uniform distribution.
Although L1, L2, and EU can all alter node weights, only EU is directly
aligned with the objective of this paper: to give a small number
of features a clear relative dominance in each node decision.

When the numerical-stability term is ignored, EU concentration is
invariant to positive rescaling of the weight vector. This property
further demonstrates that the measure characterises the relative
composition of a node\textquotesingle s decision basis across features,
rather than the overall magnitude of the weight vector.

\subsection{Scale Properties and Gradient}

Ignoring the numerical-stability term \(\epsilon\), for any scalar
\(a > 0\),

\[\rho_{ij}(aw_{i}) = \rho_{ij}(w_{i}).\]

\[C_{i}(aw_{i}) = C_{i}(w_{i}).\]

To analyse the gradient of the regulariser, first define the stabilised
absolute weight

\[a_{ij} = |w_{ij}| + \epsilon\]

Let the sum of the stabilised absolute weights be

\[S_{i} = \sum_{k = 1}^{d}{a_{ik}.}\]

Next, define the sum of squared normalised shares

\[Q_{i} = \sum_{k = 1}^{d}\rho_{ik}^{2}\]

and the dimension-normalisation constant

\[\kappa = 1 - \frac{1}{d}.\]

For \(w_{ij} \neq 0\), it further follows that

\[\frac{{\partial}C_{i}}{{\partial}w_{ij}}
=\frac{2\operatorname{sign}(w_{ij})}{\kappa S_{i}}
(\rho_{ij}-Q_{i}).\]

Because the optimiser minimises \(R_{\mathrm{EU}}\), the corresponding
regulariser gradient is

\[\frac{\partial R_{\mathrm{EU}}}{\partial w_{ij}}
=-\frac{1}{N}\frac{\partial C_i}{\partial w_{ij}}.\]

The negative sign means that minimising the regulariser increases mean
concentration. The resulting update depends on the relative feature
shares within a node and can amplify the difference between dominant
and remaining features. The full gradient still contains the task loss,
so training does not maximise concentration in isolation.

The EU-DDT parameter update satisfies

\[{\nabla}_{{\Theta}}\text{L}_{\text{EU} - \text{DDT}} = {\nabla}_{{\Theta}}\text{L}_{\text{CE}} + \lambda_{\text{EU}}{\nabla}_{{\Theta}}R_{\text{EU}}.\]

Because of the numerical-stability term, strict scale invariance does
not hold when the weights approach epsilon. This paper invokes
the property only when the weights are appreciably larger than epsilon.

\section{Evaluation Metrics}

\subsection{Predictive Performance}

Supervised learning is evaluated by test accuracy, whereas reinforcement
learning is evaluated by the success rate in independent evaluation
episodes. Supervised accuracy is defined as

\[\text{Accuracy} = \frac{1}{M}\sum_{b = 1}^{M}{\text{1}({\overset{\hat{}}{y}}_{b} = y_{b})},\]

\subsection{Node-Concentration and Degeneracy Metrics}

Concentration metrics are computed only over the set \(\mathcal{A}\) of
internal nodes not classified as collapsed. Mean active-node
concentration is

\[\overline{C}_{\mathcal{A}} =
\frac{1}{|\mathcal{A}|}\sum_{i \in \mathcal{A}}C_{i}.\]

For node \(i\), define

\[M_{i} = \max_{j}\rho_{ij}.\]

Sort the relative weights of node \(i\) in descending order and define

\[\text{Δ}_{i} = \rho_{i(1)} - \rho_{i(2)}.\]

The effective number of features converts a non-uniform distribution
into the equivalent number of uniformly used features; a lower value
indicates greater concentration:

\[N_{\text{eff},i} = \frac{1}{\sum_{j = 1}^{d}\rho_{ij}^{2}}.\]

The relative sparsity rate of node \(i\) is defined as

\[S_{i}^{(\tau)} = \frac{1}{d}\sum_{j = 1}^{d}{\text{1}(|w_{ij}| \leq \tau\max_{k}|w_{ik}|).}\]

To characterise node-weight concentration from the perspective of
distributional uncertainty, raw entropy is defined from the normalised
weight shares \(\rho_i\) as

\[H_i=-\sum_{j=1}^{d}\rho_{ij}\log\rho_{ij}.\]

Because the maximum raw entropy is \(\log d\), cross-dataset comparison
uses the normalised entropy

\[H_i^{\mathrm{norm}}=-\frac{1}{\log d}
\sum_{j=1}^{d}\rho_{ij}\log\rho_{ij},\qquad
0\leq H_i^{\mathrm{norm}}\leq 1.\]

For node \(i\), sort the relative feature weights in descending order
and define

\[k_{90,i} = min\{ k:\sum_{r = 1}^{k}{\rho_{i(r)} \geq 0.9\}}.\]

To distinguish feature concentration from collapse of the entire
node-weight vector towards zero, a node is separately classified as
collapsed using the pre-specified threshold delta:

\[\lVert w_{i}\rVert_{1} = \sum_{j = 1}^{d}{|w_{ij}| \leq \delta},\]

Mean concentration, maximum weight share, the Top-1/Top-2 margin,
effective number of features, relative sparsity rate, entropy, and k90
jointly characterise node structure. The number of collapsed nodes is
reported separately to avoid mistaking parameter degeneration for
improved interpretability.

\Needspace{6\baselineskip}\section{Supervised-Learning Experiments}

\subsection{Experimental Design}

The supervised experiments test whether EU regularisation can
substantially increase feature concentration within DDT nodes while
maintaining effective classification performance. The comparison
includes an unregularised DDT, L1-DDT, L2-DDT, EU-DDT, and a neural
network.

\subsection{Datasets and Pre-processing}

The Wine dataset contains 178 observations. Each sample is represented
by 13 input attributes, and the class label takes one of three values.
The Digits dataset contains 1,797 8 x 8 greyscale images. Each image is
flattened into 64 input features, and the class label takes one of ten
values. Each dataset is divided using a stratified 80/20 train--test
split. For Wine, features are standardised using the training-set mean
and standard deviation. For Digits, a MinMax transformation to
\([-1,1]\) is fitted on the training set. In both cases, the fitted
transformation is subsequently applied to the test set. This prevents
data leakage and improves the comparability of node weights across
features within each dataset.

\subsection{Comparison Models and Training Configuration}

Within each dataset, the four DDT variants share the same tree depth,
leaf representation, initialisation scheme, and full-batch Adam
optimiser; they differ only in the regularisation objective. A
neural-network baseline is included for comparison of predictive
performance.

All configurations are evaluated using the same five random seeds.
Tables report the mean across runs, while training curves show a single
fixed-seed trajectory for illustration only.

\subsection{Training Parameters}

The main training parameters are as follows.

\begin{longtable}[]{@{}
  >{\centering\arraybackslash}p{(\linewidth - 2\tabcolsep) * \real{0.4527}}
  >{\raggedright\arraybackslash}p{(\linewidth - 2\tabcolsep) * \real{0.4527}}@{}}
\caption{Supervised-Learning Training Parameters and Model Configuration}\\
\toprule\noalign{}
\begin{minipage}[b]{\linewidth}\centering
\textbf{Parameter}
\end{minipage} & \begin{minipage}[b]{\linewidth}\centering
\textbf{Setting}
\end{minipage} \\
\midrule\noalign{}
\endhead
\bottomrule\noalign{}
\endlastfoot
Comparison models & DDT / L1-DDT / L2-DDT / EU-DDT / Neural Network \\
Input scaling & Wine: standardisation; Digits: MinMax scaling to \([-1,1]\) \\
Tree depth & Wine: 3; Digits: 6 \\
Internal nodes / leaves & Wine: 7 / 8; Digits: 63 / 64 \\
Optimizer & Adam \\
Batching & Full batch \\
DDT learning rate & 0.03 \\
Neural-network learning rate & 0.001 \\
Neural-network architecture & MLP: 3$\times$50 hidden units, Tanh \\
Inverse temperature \(\alpha\) & 1 \\
Normalisation constant \(\epsilon\) & \(10^{-12}\) \\
Relative-sparsity threshold \(\tau\) & 0.05 \\
Collapsed-node threshold \(\delta\) & \(10^{-8}\) \\
Regularisation coefficients & \(\lambda_1=0.1\), \(\lambda_2=0.1\);
\(\lambda_{\mathrm{EU}}=5\) (Wine), \(0.1\) (Digits) \\
Wine epochs & 100 \\
Digits epochs & 1,000 \\
Reported result & Mean over five matched random seeds (curves show a
representative run) \\
\end{longtable}

The supervised-learning tables summarise mean results over five matched
random seeds and can therefore be used to assess average differences
between regularisation methods under the same initialisation conditions.
Because no hypothesis tests were conducted, this paper does not
use the term statistically significant.

\subsection{Experimental Results}

Table 6-2 summarises the mean node-weight metrics over five matched
random seeds, and Table 6-3 reports the corresponding mean test
accuracies. For Wine, the effective number of features is bounded
between 1 and 13, consistent with its 13-dimensional input
representation.

\begin{longtable}[]{@{}
  >{\centering\arraybackslash}p{(\linewidth - 12\tabcolsep) * \real{0.1184}}
  >{\raggedright\arraybackslash}p{(\linewidth - 12\tabcolsep) * \real{0.1184}}
  >{\raggedright\arraybackslash}p{(\linewidth - 12\tabcolsep) * \real{0.1499}}
  >{\raggedright\arraybackslash}p{(\linewidth - 12\tabcolsep) * \real{0.1578}}
  >{\raggedright\arraybackslash}p{(\linewidth - 12\tabcolsep) * \real{0.1342}}
  >{\raggedright\arraybackslash}p{(\linewidth - 12\tabcolsep) * \real{0.1420}}
  >{\raggedright\arraybackslash}p{(\linewidth - 12\tabcolsep) * \real{0.0947}}@{}}
\caption{Five-Seed Mean Node-Weight Statistics for Wine and Digits}\\
\toprule\noalign{}
\begin{minipage}[b]{\linewidth}\centering
\textbf{Dataset}
\end{minipage} & \begin{minipage}[b]{\linewidth}\centering
\textbf{Model}
\end{minipage} & \begin{minipage}[b]{\linewidth}\centering
\textbf{Maximum weight share $\uparrow$}
\end{minipage} & \begin{minipage}[b]{\linewidth}\centering
\textbf{Top-1/Top-2 margin $\uparrow$}
\end{minipage} & \begin{minipage}[b]{\linewidth}\centering
\textbf{Effective features $\downarrow$}
\end{minipage} & \begin{minipage}[b]{\linewidth}\centering
\textbf{Relative sparsity $\uparrow$}
\end{minipage} & \begin{minipage}[b]{\linewidth}\centering
\textbf{Mean active-node $\overline{C}_{\mathcal{A}}\uparrow$}
\end{minipage} \\
\midrule\noalign{}
\endhead
\bottomrule\noalign{}
\endlastfoot
Wine & DDT & 0.1797 & 0.0296 & 8.94 & 0.0681 & 0.0403 \\
Wine & DDT+L1 & 0.3272 & 0.0985 & 5.23 & 0.4374 & 0.1668 \\
Wine & DDT+L2 & 0.1894 & 0.0334 & 8.49 & 0.0835 & 0.0468 \\
Wine & EU-DDT & 0.7804 & 0.5906 & 1.54 & 0.8659 & 0.6962 \\
Digits & DDT & 0.0381 & 0.0022 & 45.28 & 0.0726 & 0.0069 \\
Digits & DDT+L1 & 0.1175 & 0.0360 & 30.13 & 0.3386 & 0.0554 \\
Digits & DDT+L2 & 0.0462 & 0.0059 & 43.25 & 0.1226 & 0.0095 \\
Digits & EU-DDT & 0.6950 & 0.6607 & 9.06 & 0.7989 & 0.6633 \\
\end{longtable}

Relative to DDT, EU-DDT increases the Wine maximum weight share from 0.1797
to 0.7804 and the Top-1/Top-2 margin from 0.0296 to 0.5906, reduces the
mean effective number of features from 8.94 to 1.54, and increases the
relative sparsity rate from 0.0681 to 0.8659. No collapsed internal nodes
were observed for any of the reported DDT variants under the threshold
\(\delta=10^{-8}\): each Wine variant had 0/35 collapsed nodes across
five seeds, and each Digits variant had 0/315 collapsed nodes. Hence,
\(\overline{C}_{\mathcal{A}}=\overline{C}\) for all reported supervised
runs.
Mean concentration correspondingly increases from 0.0403 to 0.6962.

The Digits results follow the same direction. Relative to DDT, EU-DDT
increases the maximum weight share from 0.0381 to 0.6950 and the
Top-1/Top-2 margin from 0.0022 to 0.6607, reduces the mean effective
number of features from 45.28 to 9.06, and increases the relative
sparsity rate from 0.0726 to 0.7989. The concentration gain is therefore
not specific to the 13-dimensional Wine inputs; it also holds for the
64-dimensional image features. Mean concentration increases from 0.0069
to 0.6633. Because \(C_i\leq M_i\) at every node, the recomputed
five-seed aggregates also satisfy \(\overline C\leq\overline M\) for
every reported model-dataset configuration.

EU-DDT attains mean accuracies of 93.9\% on Wine and 94.6\% on Digits,
compared with 100.0\% and 92.3\% for the unregularised DDT. The effective
number of features decreases by 82.8\% and 80.0\%, respectively, and the
dominant-weight metrics increase substantially on both datasets. The
selected EU regularisation strengths therefore provide a favourable
performance-concentration configuration: the models still perform the
classification tasks effectively while the basis of each split becomes
more concentrated.

\begin{longtable}[]{@{}
  >{\centering\arraybackslash}p{(\linewidth - 10\tabcolsep) * \real{0.2444}}
  >{\raggedright\arraybackslash}p{(\linewidth - 10\tabcolsep) * \real{0.1340}}
  >{\raggedright\arraybackslash}p{(\linewidth - 10\tabcolsep) * \real{0.1340}}
  >{\raggedright\arraybackslash}p{(\linewidth - 10\tabcolsep) * \real{0.1340}}
  >{\raggedright\arraybackslash}p{(\linewidth - 10\tabcolsep) * \real{0.1340}}
  >{\raggedright\arraybackslash}p{(\linewidth - 10\tabcolsep) * \real{0.2128}}@{}}
\caption{Five-Seed Mean Test Accuracy on Wine and Digits}\\
\toprule\noalign{}
\begin{minipage}[b]{\linewidth}\centering
\textbf{Dataset}
\end{minipage} & \begin{minipage}[b]{\linewidth}\centering
\textbf{DDT}
\end{minipage} & \begin{minipage}[b]{\linewidth}\centering
\textbf{DDT+L1}
\end{minipage} & \begin{minipage}[b]{\linewidth}\centering
\textbf{DDT+L2}
\end{minipage} & \begin{minipage}[b]{\linewidth}\centering
\textbf{EU-DDT}
\end{minipage} & \begin{minipage}[b]{\linewidth}\centering
\textbf{Neural Network}
\end{minipage} \\
\midrule\noalign{}
\endhead
\bottomrule\noalign{}
\endlastfoot
Wine five-seed mean accuracy & 1.000 & 1.000 & 1.000 & 0.939 & 1.000 \\
Digits five-seed mean accuracy & 0.923 & 0.949 & 0.960 & 0.946 &
0.969 \\
\end{longtable}

\subsection{Weight and Training Curves}

Figure 6-1 displays the absolute feature weights of selected internal
nodes on Digits; the high-weight pixels of EU-DDT are visibly
concentrated in fewer input dimensions. Figure 6-2 presents test-accuracy
and raw node-weight-entropy curves for representative Wine and Digits
seeds to illustrate the within-dataset optimisation process. Because
\(d\) is fixed within each panel, these raw entropy trajectories are not
compared numerically across datasets. Tables 6-2 and 6-3 report the
five-seed aggregate results.

\begin{figure}[htbp]
\centering
\includegraphics[width=0.48\linewidth]{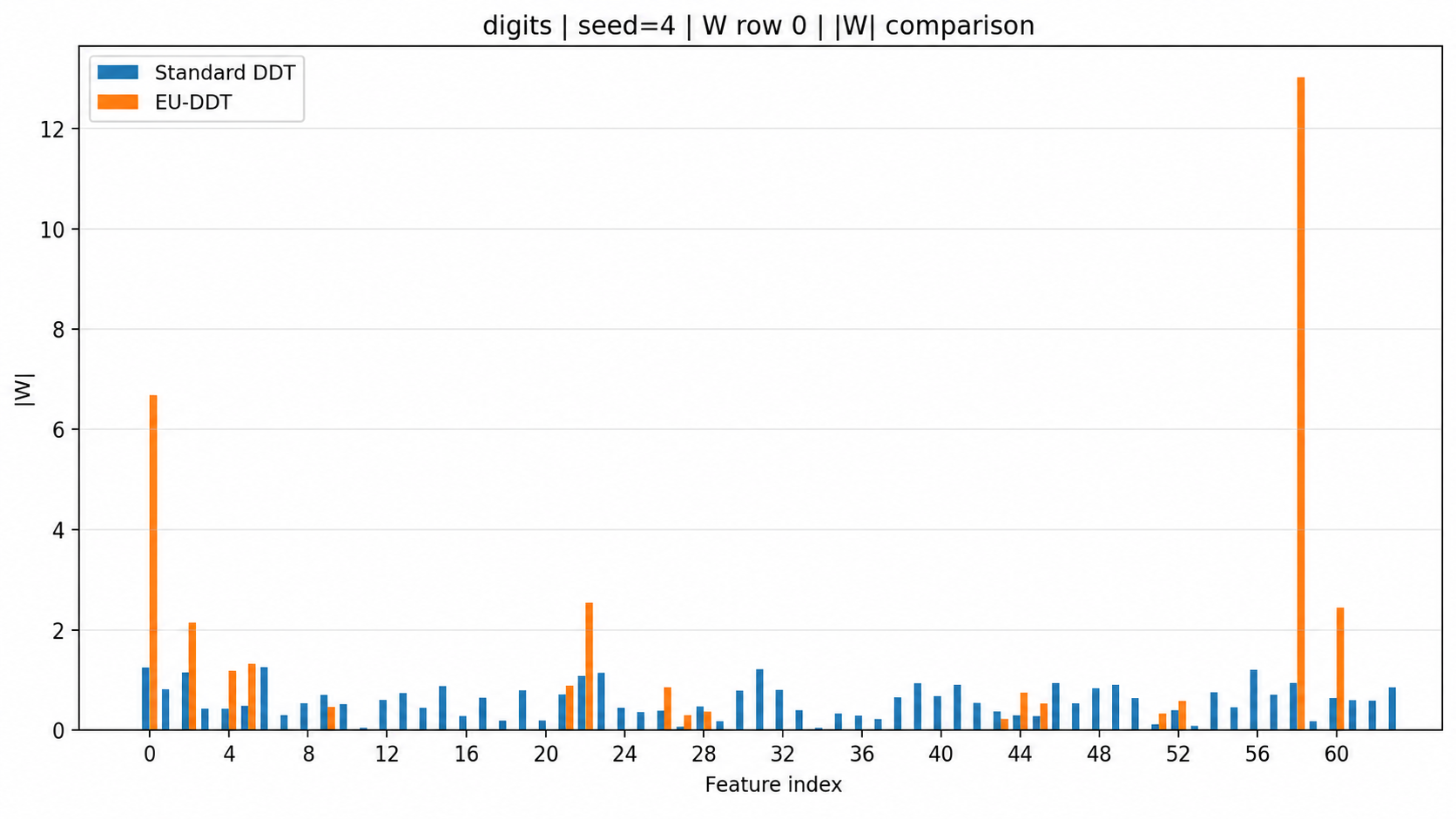}\hfill
\includegraphics[width=0.48\linewidth]{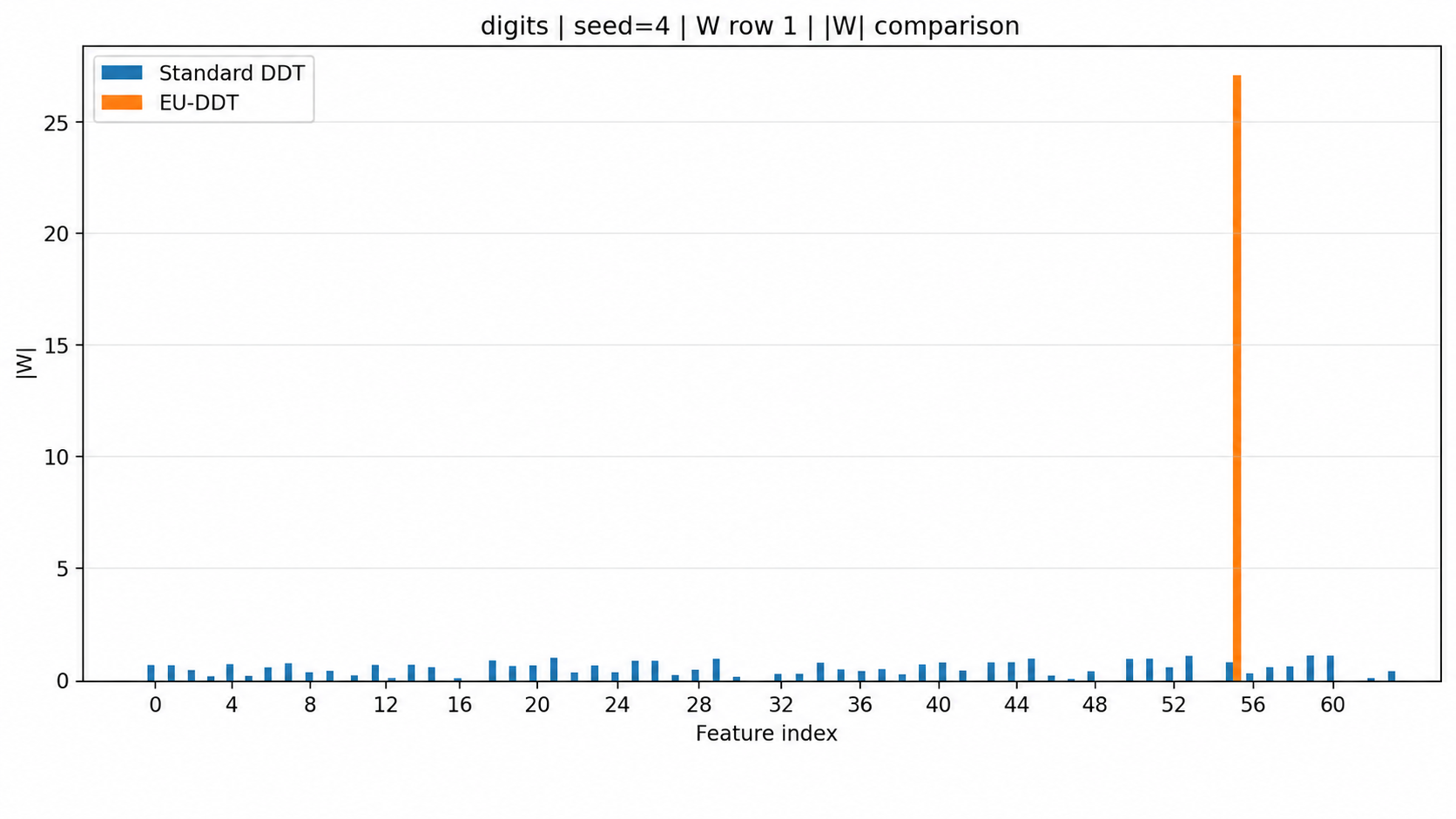}
\par\smallskip
\includegraphics[width=0.48\linewidth]{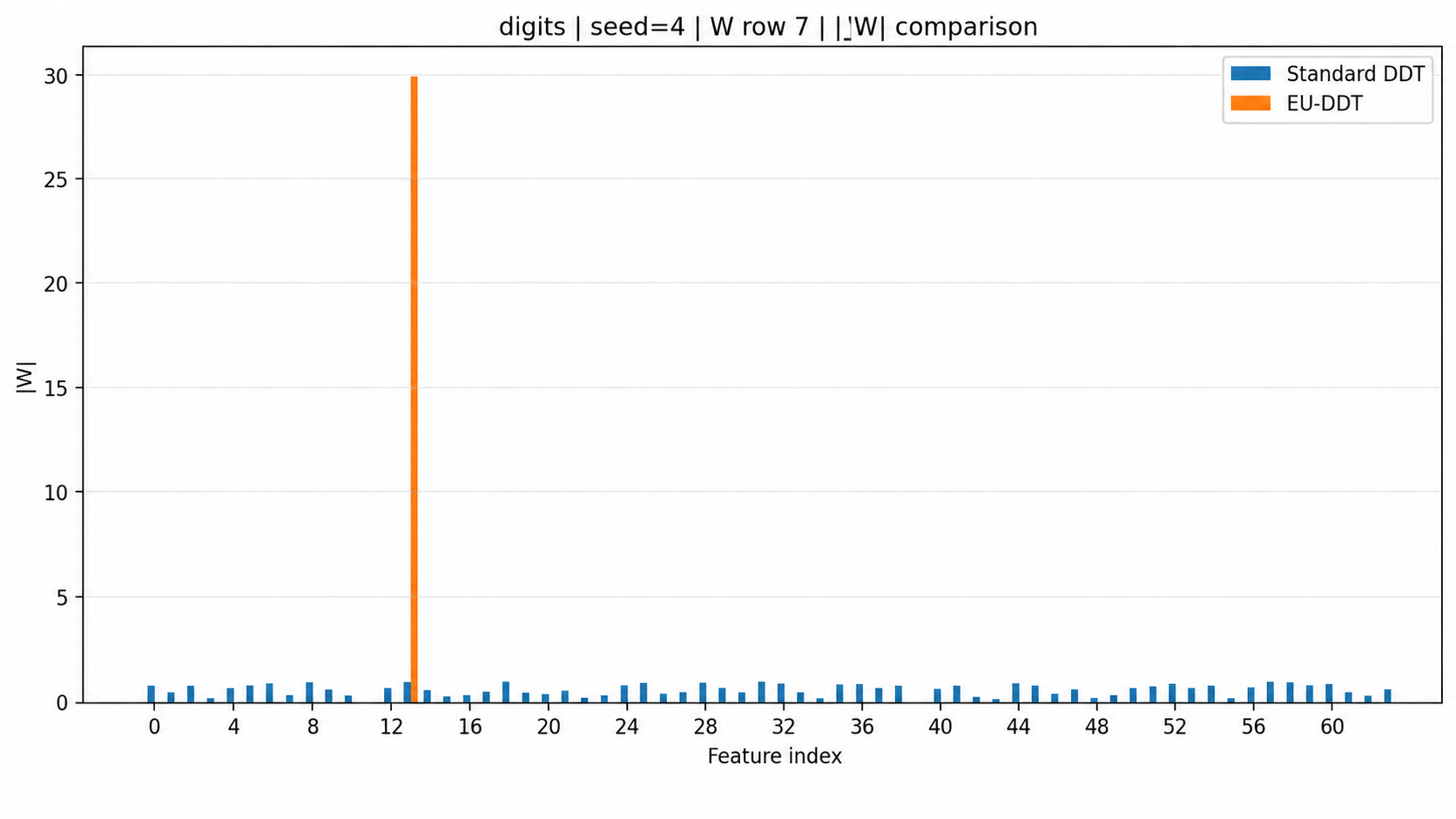}\hfill
\includegraphics[width=0.48\linewidth]{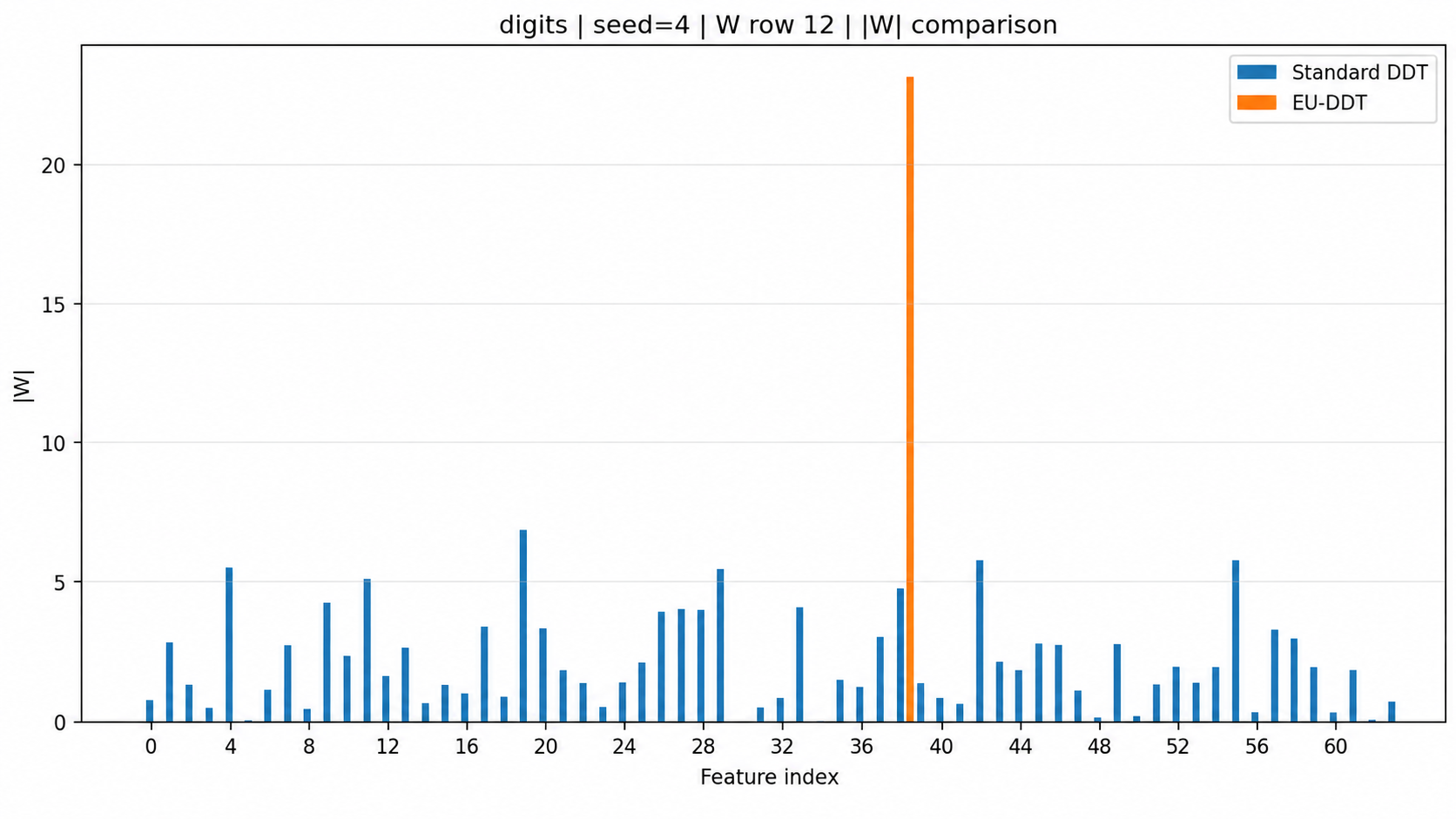}
\caption{Comparison of Absolute Feature Weights at Selected Internal Nodes on Digits.}
\end{figure}

\begin{figure}[htbp]
\centering
\includegraphics[width=0.48\linewidth]{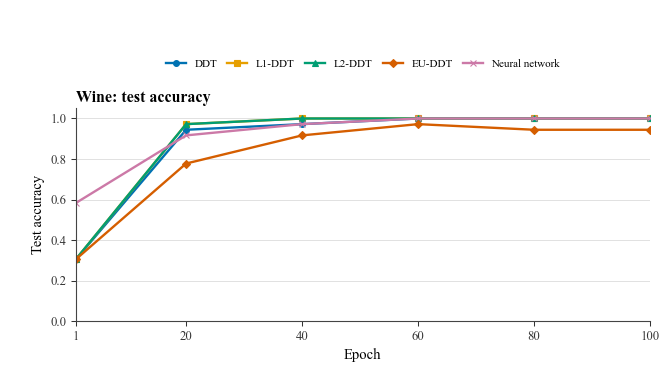}\hfill
\includegraphics[width=0.48\linewidth]{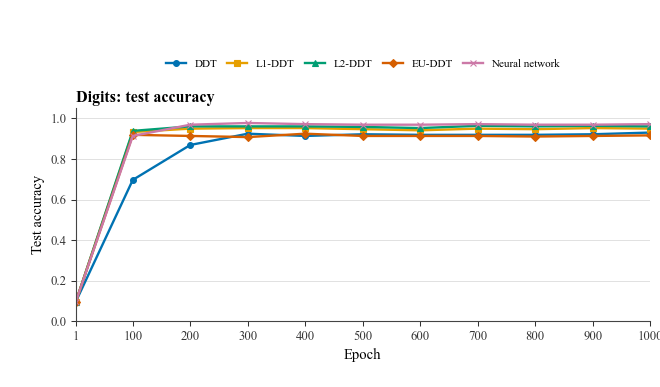}
\par\smallskip
\includegraphics[width=0.48\linewidth]{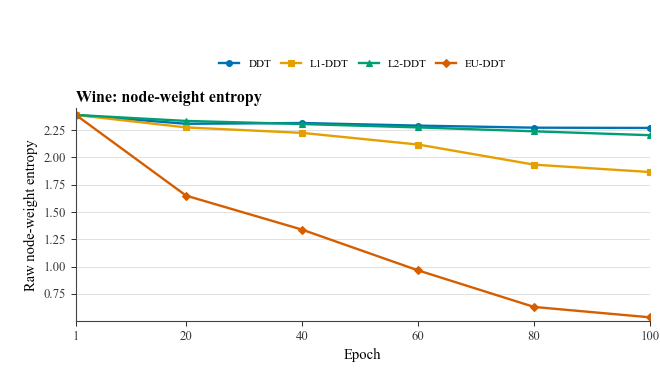}\hfill
\includegraphics[width=0.48\linewidth]{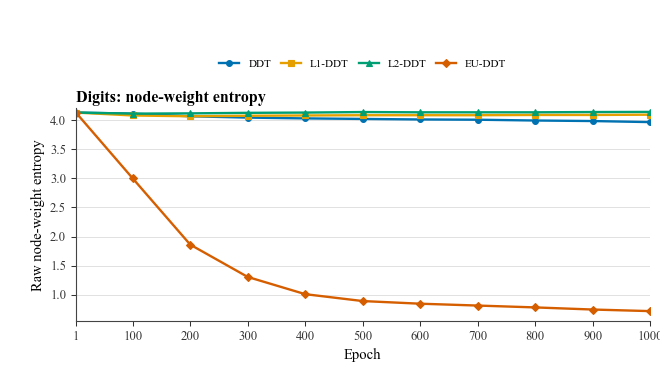}
\caption{Test Accuracy and Node-Weight Entropy Curves on Wine and Digits.}
\end{figure}

\subsection{Interpretation of the Results}

The claim of improved interpretability is supported by three mutually
reinforcing levels of evidence. First, the fixed tree topology and
explicit root-to-leaf paths make the overall decision structure
traceable. Second, on both Wine and Digits, EU-DDT increases the maximum
weight share, Top-1/Top-2 margin, and relative sparsity rate, while
reducing the effective number of features by 82.8\% on Wine and 80.0\%
on Digits. The
basis of each node decision therefore shifts from diffuse participation
by many features to dominance by a small number of features, yielding
clearer feature attribution at the node level. Third, the model retains
a mean test accuracy above 93\% on both datasets, showing that the
concentrated structure belongs to a model that remains predictively
effective rather than reflecting artificial simplicity produced by
parameter collapse.

Under this operational definition, the interpretability improvement of
EU-DDT has two components: the decision path itself is inspectable, and
each node along that path is determined by fewer and more dominant input
features. This provides a lower-dimensional and more directly
inspectable node-level attribution structure. The paper therefore
concludes that intrinsic structural interpretability is improved. A
human-subject study could provide subsequent external validation of this
structural evidence, but is not required for the direct measurements of
node-wise dependency dimension and dominant-feature strength reported
here.

\Needspace{6\baselineskip}\section{Application of EU-DDT to Reinforcement Learning}

\subsection{Research Design}

This section directly parameterises a PPO actor with EU-DDT to test
whether the node-concentration mechanism developed in supervised
learning transfers to policy learning. Independent task success rates
and the feature-concentration structure of the actor nodes are evaluated
jointly.

The EU-DDT actor maps the current state directly to an action
distribution, and PPO optimises the tree policy through the clipped
policy objective. The MLP critic is used only for value estimation and
advantage construction. Decisions at deployment therefore remain
governed by the inspectable EU-DDT tree paths and dominant state
features.

\subsection{PPO and the EU-DDT Policy}

PPO limits the magnitude of policy updates by comparing the probability
assigned to an executed action by the current policy with that assigned
by the pre-update policy \citep{schulman2017ppo}. Define the probability ratio

\[r_{t}(\theta) = \frac{\pi_{\theta}(a_{t} \mid s_{t})}{\pi_{\theta_{\text{old}}}(a_{t} \mid s_{t})}.\]

The clipped objective is

\[L_{\text{CLIP}}(\theta) = \text{E}_{t}\lbrack min(r_{t}(\theta){\overset{\hat{}}{A}}_{t},\ clip(r_{t}(\theta),1 - \epsilon,1 + \epsilon){\overset{\hat{}}{A}}_{t})\rbrack,\]

where \({\overset{\hat{}}{A}}_{t}\) is the advantage estimate and
\(\epsilon\) is the clipping parameter. This objective limits excessive
changes in the probability ratio between the new and old policies,
thereby reducing the magnitude of an individual policy update.

\subsection{EU-DDT Actor and Training Objective}

In supervised learning, the DDT input is the feature vector \(x\). In
reinforcement learning, the same architecture instead receives the
current environment state:

\[g_{i}(s_{t}) = \sigma(\alpha(w_{i}^{{\top}}s_{t} - c_{i})).\]

The policy distribution is

\[\pi_{\theta}(a \mid s_{t}) = Softmax(\sum_{\ell = 1}^{L}{\pi_{\ell}(s_{t})a_{\ell})}.\]

This architecture can be trained by standard backpropagation and allows
the state variables on which each internal node primarily depends to be
inspected.

\subsection{Actor and Critic Losses}

The main code implements the PPO actor objective in minimisation form
and includes policy entropy, L1, L2, and EU concentration terms. Let
\(E\) be the number of epochs per PPO update, \(n\) the number of
samples, and \(m\) the minibatch size. The actor loss for one minibatch,
consistent with the implementation, is

\[\text{L}_{\text{actor}} = - L_{\text{CLIP}} - c_{H}H(\pi_{\theta}) + \eta_{R}(\lambda_{1}R_{\text{L}1} + \lambda_{2}R_{\text{L}2} + \lambda_{\text{EU}}R_{\text{EU}}).\]

Here, \(c_{H}\) is the policy-entropy coefficient at the current update,
and \(\eta_{R} = \lbrack E\lceil n/m\rceil\rbrack^{- 1}\) is
regulariser\_scale in the code. This scaling prevents the same
regularisation term from being accumulated repeatedly across the
multiple epochs and minibatches of a single PPO update. For each
ablation group, only the corresponding \(\lambda_{1}\), \(\lambda_{2}\),
or \(\lambda_{\text{EU}}\) is active.

\subsection{Value Network}

The MLP critic uses GAE to construct return targets and minimises

\[\text{L}_{\text{critic}} = c_{V}\ \text{E}_{t}\lbrack(V_{\phi}(s_{t}) - {\overset{\hat{}}{R}}_{t})^{2}\rbrack,\ c_{V} = 0.5.\]

EU-DDT directly produces the action distribution, whereas the critic is
used only for value estimation and advantage construction. The results
therefore focus on the interpretability of the policy representation.

\subsection{Experimental Configuration}

The ablation study compares an unregularised PPO-DDT with L1- and
L2-regularised variants, for which \(\lambda_{1} = \lambda_{2} = 1\), and
with PPO-EU-DDT. The Euclidean concentration coefficient is set to
\(\lambda_{\mathrm{EU}} = 0.3\) for CartPole-v1 and
\(\lambda_{\mathrm{EU}} = 1\) for MiniPacman. All variants employ the
same DDT actor architecture, while the environment, training budget,
and PPO hyperparameters are held fixed.

For CartPole-v1, success is the proportion of evaluation episodes with a
return of at least 475. For MiniPacman, success is the proportion of
evaluation episodes in which the map is cleared. The main configuration
is as follows.

\begin{longtable}[]{@{}
  >{\centering\arraybackslash}p{(\linewidth - 4\tabcolsep) * \real{0.3644}}
  >{\raggedright\arraybackslash}p{(\linewidth - 4\tabcolsep) * \real{0.3178}}
  >{\raggedright\arraybackslash}p{(\linewidth - 4\tabcolsep) * \real{0.3178}}@{}}
\caption{Reinforcement-Learning Experimental Configuration}\\
\toprule\noalign{}
\begin{minipage}[b]{\linewidth}\centering
\textbf{Parameter}
\end{minipage} & \begin{minipage}[b]{\linewidth}\centering
\textbf{CartPole-v1}
\end{minipage} & \begin{minipage}[b]{\linewidth}\centering
\textbf{MiniPacman}
\end{minipage} \\
\midrule\noalign{}
\endhead
\bottomrule\noalign{}
\endlastfoot
Representative seed & 3 & 1 \\
State dimension & 4 & 14 \\
Action dimension & 2 & 4 \\
Tree depth & 3 & 4 \\
Internal nodes / leaves & 7 / 8 & 15 / 16 \\
PPO updates & 60 & 60 \\
PPO optimisation epochs/update & 4 & 4 \\
Environment interactions & 61,440 & 138,240 \\
Critic & MLP: 3$\times$64, Tanh & MLP: 3$\times$64, Tanh \\
\end{longtable}

\subsection{Experimental Results}

Each environment was run under two independent random seeds, and both
repetitions showed that EU-DDT increased node concentration while
maintaining effective task performance. Table 7-2 and Figure 7-1 report
representative runs for which the curves and node metrics correspond
directly: CartPole-v1 uses seed = 3 and MiniPacman uses seed = 1; both
tasks are trained for 60 PPO updates.

\begin{table}[htbp]
\centering
\caption{Reinforcement-Learning Results for Representative Random Seeds}
\label{tab:rl-results}
\small
\setlength{\tabcolsep}{7pt}
\begin{tabular}{@{}lrrrr@{}}
\toprule
\multicolumn{5}{@{}l}{\textbf{CartPole-v1}} \\
Metric & DDT & L1 & L2 & EU \\
\midrule
Final success rate $\uparrow$ & 100\% & 90\% & 46.7\% & 100\% \\
Concentration $\uparrow$ & 0.1272 & 0.3153 & 0.0798 & 0.7179 \\
Maximum share $\uparrow$ & 0.4661 & 0.6031 & 0.4114 & 0.8220 \\
Top-1/Top-2 margin $\uparrow$ & 0.1648 & 0.3562 & 0.1272 & 0.7236 \\
k90 $\downarrow$ & 3.29 & 2.86 & 3.71 & 1.71 \\
Effective features $\downarrow$ & 2.94 & 2.28 & 3.31 & 1.62 \\
Relative sparsity $\uparrow$ & 0.0000 & 0.1071 & 0.0000 & 0.5357 \\
\midrule
\multicolumn{5}{@{}l}{\textbf{MiniPacman}} \\
Metric & DDT & L1 & L2 & EU \\
\midrule
Final success rate $\uparrow$ & 100\% & 100\% & 100\% & 100\% \\
Concentration $\uparrow$ & 0.0437 & 0.1921 & 0.0658 & 0.7237 \\
Maximum share $\uparrow$ & 0.1793 & 0.3391 & 0.2383 & 0.8205 \\
Top-1/Top-2 margin $\uparrow$ & 0.0340 & 0.1909 & 0.0892 & 0.7623 \\
k90 $\downarrow$ & 10.00 & 7.93 & 9.47 & 2.93 \\
Effective features $\downarrow$ & 9.46 & 6.90 & 8.23 & 1.96 \\
Relative sparsity $\uparrow$ & 0.0810 & 0.2714 & 0.1286 & 0.7667 \\
\bottomrule
\end{tabular}
\end{table}

\begin{figure}[htbp]
\centering
\includegraphics[width=0.48\linewidth]{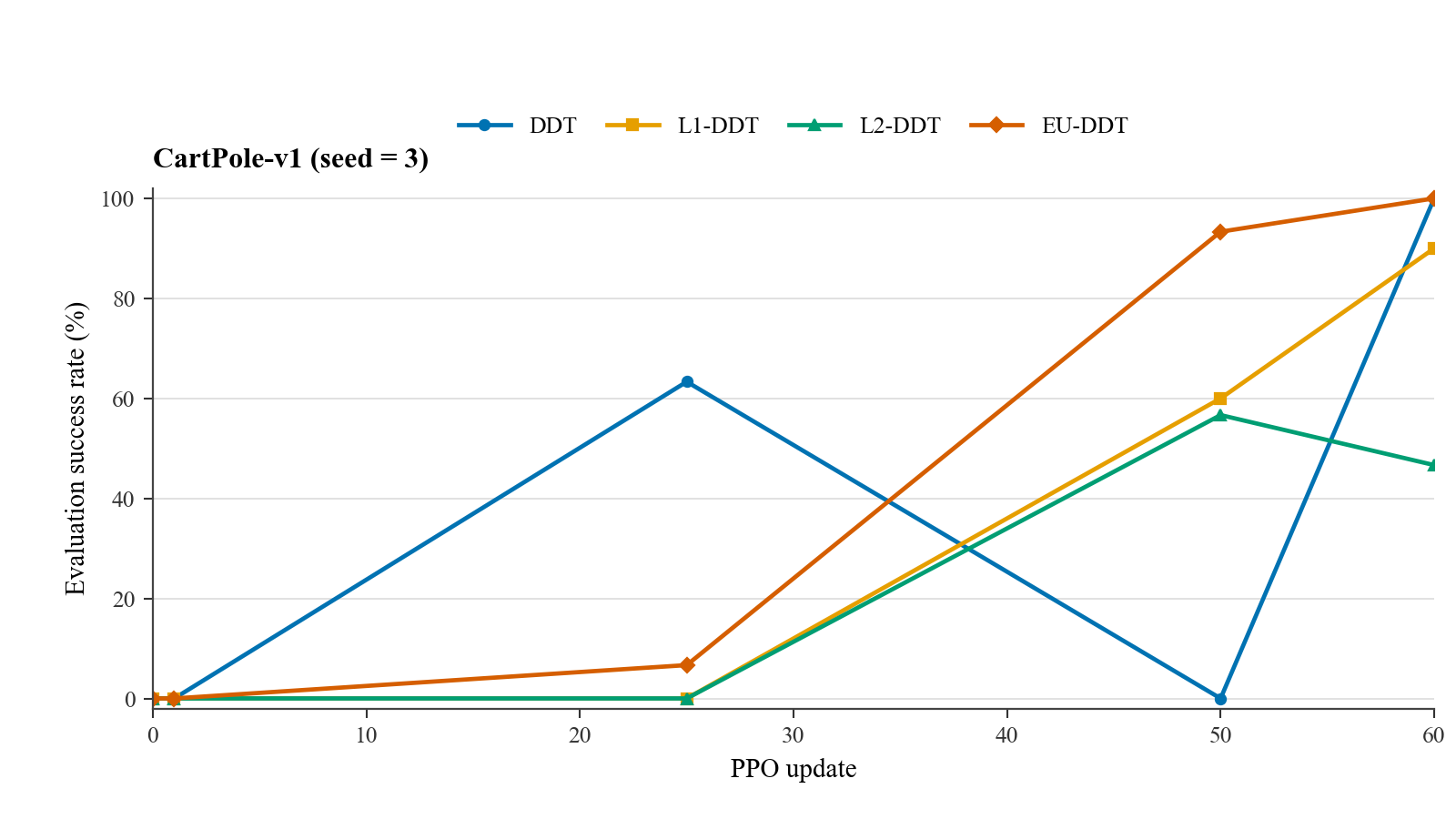}\hfill
\includegraphics[width=0.48\linewidth]{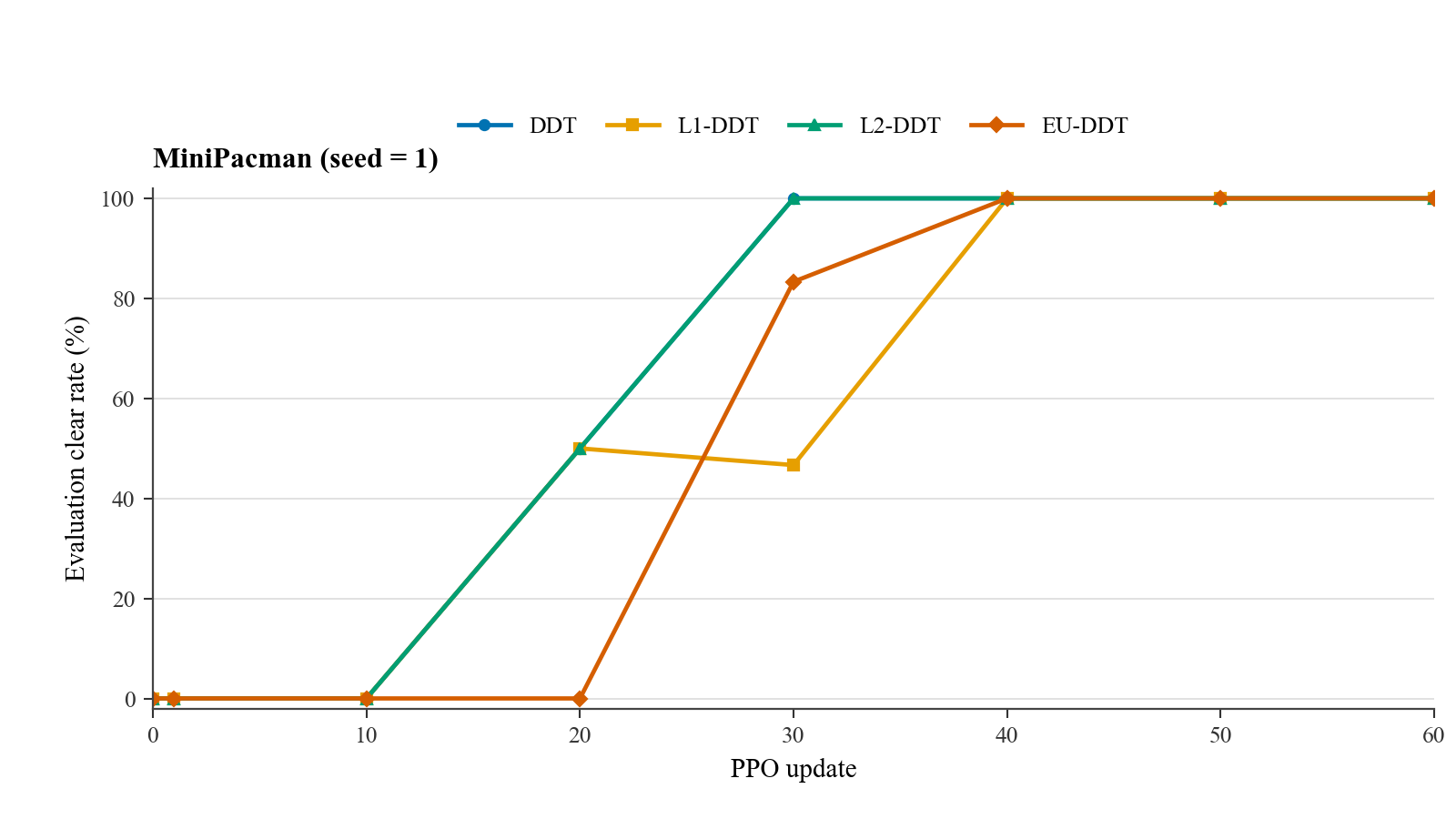}
\par\smallskip
\includegraphics[width=0.48\linewidth]{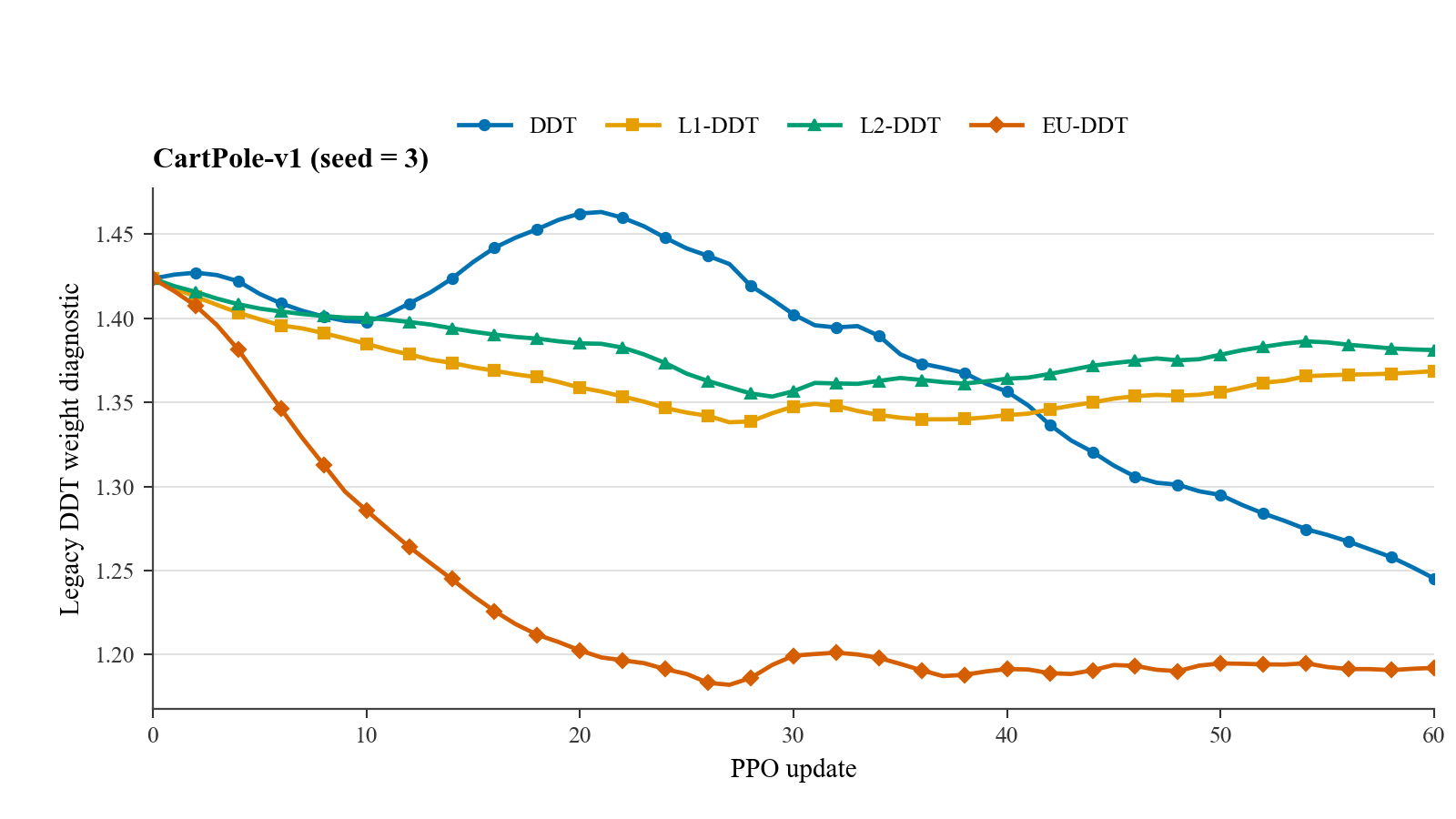}\hfill
\includegraphics[width=0.48\linewidth]{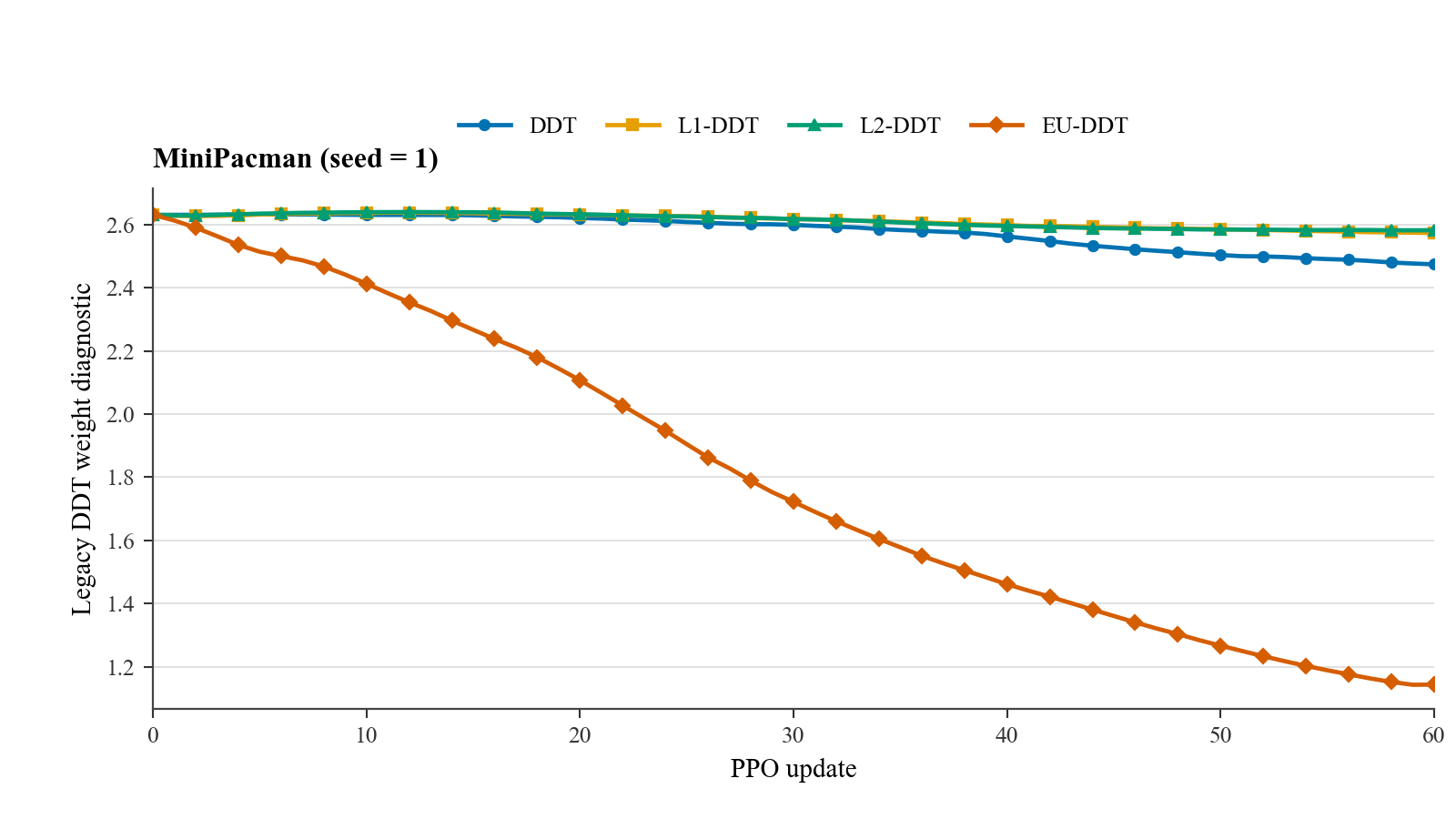}
\caption{Independent-Evaluation Success Rate and Legacy DDT Weight-Diagnostic Curves for Representative Runs on CartPole-v1 and MiniPacman.}
\end{figure}

The lower panels report the project's legacy
\texttt{DDT.entroy\_loss()} diagnostic, which combines a mean
absolute-weight term with an entropy term based on softmax-transformed
absolute weights. It is not the raw share entropy
\(H_i=-\sum_j\rho_{ij}\log\rho_{ij}\) defined in Section 5 and is
therefore not bounded by \(\log d\). The concentration metrics in Table
7-2 are computed separately from normalised absolute-weight shares.

At update 60 on CartPole-v1, the success rates of DDT, L1-DDT, L2-DDT,
and EU-DDT are 100\%, 90\%, 46.7\%, and 100\%, respectively. EU-DDT
increases mean concentration from 0.1272 to 0.7179 and the maximum
weight share from 0.4661 to 0.8220, while reducing k90 from 3.29 to
1.71.

On MiniPacman, all four policies achieve a map-clearance rate of 100\%
at update 60. EU-DDT increases mean concentration from 0.0437 to 0.7237
and the maximum weight share from 0.1793 to 0.8205, while reducing k90
from 10.00 to 2.93.

The two-seed repetitions in both environments exhibit the same central
pattern: EU-DDT maintains effective final task performance while
producing higher node concentration than the unregularised DDT and
reducing the number of features required to account for 90\% of the
cumulative weight. In the representative runs reported in Table 7-2,
EU-DDT achieves a final success rate of 100\% in both environments. The
two-seed evidence therefore supports a consistent empirical balance
between policy-performance preservation and simplification of the
node-level explanatory structure. Because each environment uses only two
seeds, the evidence is used to assess directional consistency rather
than to infer statistical significance.

\Needspace{6\baselineskip}\section{Discussion}

\subsection{Node Concentration and Predictive Performance}

The five-seed means on Wine and Digits show that EU regularisation
reduces the effective number of features by 82.8\% and 80.0\%,
respectively, in tasks with different input dimensionalities, while increasing the
dominant-weight share, Top-1/Top-2 margin, and relative sparsity rate.
Together with the clearer dominant input dimensions visible in the
weight visualisations, these complementary
metrics demonstrate simpler feature attribution at the node level. At
the same time, EU-DDT retains a mean accuracy above 93\% on both
datasets and maintains classification performance close to the
baselines. The principal finding is therefore not a simple exchange of
accuracy for interpretability. Rather, EU-DDT obtains a much larger
improvement in node concentration for a comparatively small difference
in performance, producing an effective balance between predictive
performance and intrinsic structural interpretability.

\subsection{Relationship to Existing Methods}

EU-DDT can be distinguished from existing methods by its optimisation
target. Previous work has primarily developed differentiable splits,
representation learning, adaptive tree structures, or discretisation
strategies \citep{kontschieder2015dndf,yang2018dndt,tanno2019ant}, and has applied DDTs to policy-gradient optimisation \citep{silva2020ddtrl}. This paper does not change the tree topology or
predictive function. Instead, it introduces into the probabilistic soft
tree a Euclidean regularisation objective that directly targets relative
feature concentration within each node. L1 and L2 constrain parameter
magnitude, whereas EU optimises the position of the normalised weights
on the probability simplex relative to the uniform distribution. EU-DDT
yields a lower mean effective-feature count and higher dominant-weight
metrics than L1-DDT and L2-DDT under the reported settings, indicating
that direct optimisation of relative concentration is better aligned
with the interpretability objective of this paper.

\subsection{Methodological Position of the Continuous Concentration
Mechanism}

NODE uses entmax to produce sparse feature selection with exact zeros,
whereas ICCT learns few-feature or single-feature decisions through
differentiable crispification \citep{popov2020node,paleja2022icct}.
These methods seek sparse or crisp split representations. EU-DDT
addresses a different but complementary problem: it gives a small number
of features clear dominance within an oblique soft split without
applying hard argmax or altering the original predictive function. This
continuous mechanism avoids the potential fidelity discrepancy
introduced by first training a soft model and then discretising it into
a separate explanation model. The object being inspected is therefore
the original EU-DDT that is actually used for prediction or
decision-making.

\subsection{Preservation of Task Performance and Gains in Node
Concentration}

The joint EU-DDT objective combines classification or policy-learning
performance with the node-concentration objective in a single
optimisation process:

\[\text{L} = \text{L}_{\text{task}} + \lambda_{\text{EU}}R_{\text{EU}}.\]

The selected EU concentration coefficient provides an effective,
outcome-oriented configuration in both task settings. In supervised
learning, test accuracy remains above 93\% while the effective number of
features decreases by 82.8\% on Wine and 80.0\% on Digits. In the two-seed
reinforcement-learning repetitions, EU-DDT maintains effective task
performance and exhibits consistent gains in node concentration; in the
representative runs, final success rates are 100\% and mean
concentration exceeds 0.71 in both environments. EU regularisation
therefore does not obtain superficial simplicity at the cost of task
failure. It substantially increases the concentration and inspectability
of the node-level decision basis while the model remains capable of
effective prediction or control.

\subsection{Limitations}

The supervised-learning experiments use five random seeds and show
consistent average trends across the compared methods. Nevertheless,
additional repetitions would provide more reliable estimates of
variance, while evaluation on a broader range of datasets would
strengthen evidence of generalisability. The reinforcement-learning
experiments use only two random seeds per environment. Although both
runs show the same qualitative pattern in task performance and node
concentration, this is insufficient for reliable statistical inference,
and further repetitions are required.

The interpretation of node weights also depends on appropriate feature
scaling. Each dataset therefore uses a fixed scaling strategy across all
compared models---standardisation for Wine and MinMax scaling to
\([-1,1]\) for Digits---with parameters fitted on the training data only
to avoid data leakage. In addition, the interpretability analysis in this paper
is based primarily on structural properties of the learned trees and
node-weight concentration. Future work could complement these measures
with human-subject evaluation or assess the fidelity of simplified
single-feature rules derived from the learned nodes.

\Needspace{6\baselineskip}\section{Conclusions and Future Work}

\subsection{Main Conclusions}

This paper proposes EU-DDT, which regularises the relative
feature-weight distribution within each DDT node using a normalised
squared Euclidean concentration objective. The method retains
sigmoid-based soft routing and end-to-end differentiability while
encouraging decisions to rely on a small number of dominant features.
Unlike L1 and L2 regularisation, EU regularisation acts on relative
weight concentration rather than parameter magnitude.

Across five random seeds, the supervised-learning experiments show that
EU-DDT reduces node-level feature complexity on both Wine and Digits. The
mean effective number of features decreases by 82.8\% and 80.0\%,
respectively, while leading-feature dominance increases substantially
and mean test accuracy remains above 93\%.

The reinforcement-learning experiments show the same qualitative trend
in CartPole-v1 and MiniPacman. Across two random seeds per environment,
EU-DDT increases node concentration while preserving successful task
performance; the reported evaluations achieve a final success rate of
100\%, with mean node concentration above 0.71.

Overall, the results support EU regularisation as a continuous,
differentiable mechanism for reducing DDT node complexity while retaining
effective predictive and policy performance.

\subsection{Answers to the Research Questions}

For RQ1, the five matched random-seed results show that EU
regularisation substantially increases the maximum node-weight share and
the Top-1/Top-2 margin on both Wine and Digits, while reducing the mean
effective number of features by 82.8\% and 80.0\%, respectively. This evidence
supports the conclusion that normalised squared Euclidean concentration
encourages DDT internal nodes to form weight representations dominated
by a small number of features.

For RQ2, the five-seed mean test accuracies of EU-DDT are 93.9\% on Wine
and 94.6\% on Digits, compared with 100.0\% and 92.3\% for the
corresponding unregularised DDTs. While retaining high classification
accuracy, EU-DDT produces markedly larger gains in node concentration
than L1-DDT and L2-DDT. The corresponding neural-network accuracies are
100.0\% on Wine and 96.9\% on Digits, providing an external
predictive-performance reference. Under the experimental settings used in this
paper, EU-DDT therefore achieves an effective balance between
predictive performance and intrinsic structural interpretability.

For RQ3, two independent random-seed runs in each environment show that
EU-DDT can parameterise the PPO actor and output the action distribution
directly, while increasing actor-node concentration and maintaining
effective task performance. In the representative runs, CartPole-v1
(seed = 3) and MiniPacman (seed = 1) both achieve a final evaluation
success rate of 100\%, with mean node concentrations of 0.7179 and
0.7237, respectively. The two-seed results support the extension of the
EU concentration mechanism from supervised classification to sequential
decision-making and provide preliminary evidence of repeatability.
Stronger cross-seed statistical inference requires a larger number of
repetitions.

\subsection{Future Work}

Future work will increase the number of reinforcement-learning random
seeds and report means, standard deviations, and confidence intervals in
additional environments, extending the present two-seed
directional-consistency assessment into a formal statistical evaluation.
It could also systematically trace the performance-concentration
frontier associated with the concentration coefficient and add
human-subject studies to quantify how tree paths and dominant features
improve comprehension efficiency. For applications requiring symbolic
rules, EU-DDT could additionally serve as the starting point for a
high-fidelity simplification process, with behavioural consistency
evaluated after top-feature crispification.

\clearpage
\bibliographystyle{plainnat}
\bibliography{references}

\end{document}